\documentclass[10pt,twocolumn,letterpaper]{article}
\usepackage{cvpr}              

\usepackage{graphicx} 
\usepackage{tikz}
\usetikzlibrary{positioning, arrows.meta, fit, backgrounds}
\usepackage{float}
\usepackage{caption}
\definecolor{cvprblue}{rgb}{0.21,0.49,0.74}
\usepackage[pagebackref,breaklinks,colorlinks,allcolors=cvprblue]{hyperref}

\usepackage[hyphens]{url}          
\usepackage{multirow}

\hypersetup{colorlinks=true,linkcolor=red,citecolor=blue,urlcolor=magenta}
\newcommand{\smallurl}[1]{\footnotesize\url{#1}}

\usepackage{pifont}

\def\paperID{*****} 
\def\confName{CVPR}
\def\confYear{2026}

\title{Robust Multi-Source Covid-19 Detection in CT Images}

\author{
Asmita Yuki Pritha\textsuperscript{1}\thanks{Co-first Authors}\quad
Jason Xu\textsuperscript{2}$^*$ \quad
Daniel Ding\textsuperscript{2}$^*$ \quad
Justin Li\textsuperscript{2}$^*$\quad
Aryana Hou \textsuperscript{3}$^*$\quad
Xin Wang \textsuperscript{4}\quad \\
Shu Hu\textsuperscript{5}\thanks{Corresponding author.}\\
\textsuperscript{1} Capstone School Dhaka, Bangladesh {\tt\small  asmitayukipritha@gmail.com}\\
\textsuperscript{2} Carmel High School, Carmel, Indiana, USA {\tt\small  \{jason013229, danielding56, justinyli2018\}@gmail.com}\\
\textsuperscript{3} Clarkstown High School South, West Nyack, New York, USA  {\tt\small  aryanahou@gmail.com}
\\
\textsuperscript{4} University at Albany, State University of New York, Albany, New York, USA  {\tt\small  xwang56@albany.edu}
\\
\textsuperscript{5}Purdue University, West Lafayette, Indiana, USA
{\tt\small hu968@purdue.edu}
}
\begin{document}
\maketitle

\begin{abstract}
Deep learning models for COVID-19 detection from chest CT scans generally perform well when the
training and test data originate from the same institution, but they often struggle when scans are drawn
from multiple centres with differing scanners, imaging protocols, and patient populations. One key
reason is that existing methods treat COVID-19 classification as the sole training objective, without
accounting for the data source of each scan. As a result, the learned representations tend to be biased
toward centres that contribute more training data. To address this, we propose a multi-task learning
approach in which the model is trained to predict both the COVID-19 diagnosis and the originating data
centre. The two tasks share an EfficientNet-B7 backbone, which encourages the feature
extractor to learn representations that hold across all four participating centres. Since the training data
is not evenly distributed across sources, we apply a logit-adjusted cross-entropy loss~\cite{menon21}
to the source classification head to prevent underrepresented centres from being overlooked. Our
preprocessing follows the SSFL framework with KDS~\cite{hsu24}, selecting eight representative slices
per scan.
Our method achieves an F1 score of 0.9098 and an AUC-ROC of
0.9647 on a validation set of 308 scans. The code is publicly available at \url{https://github.com/Purdue-M2/-multisource-covid-ct}.
\end{abstract}

\section{Introduction}
\begin{figure}[t]
\centering
\includegraphics[width=0.95\columnwidth]{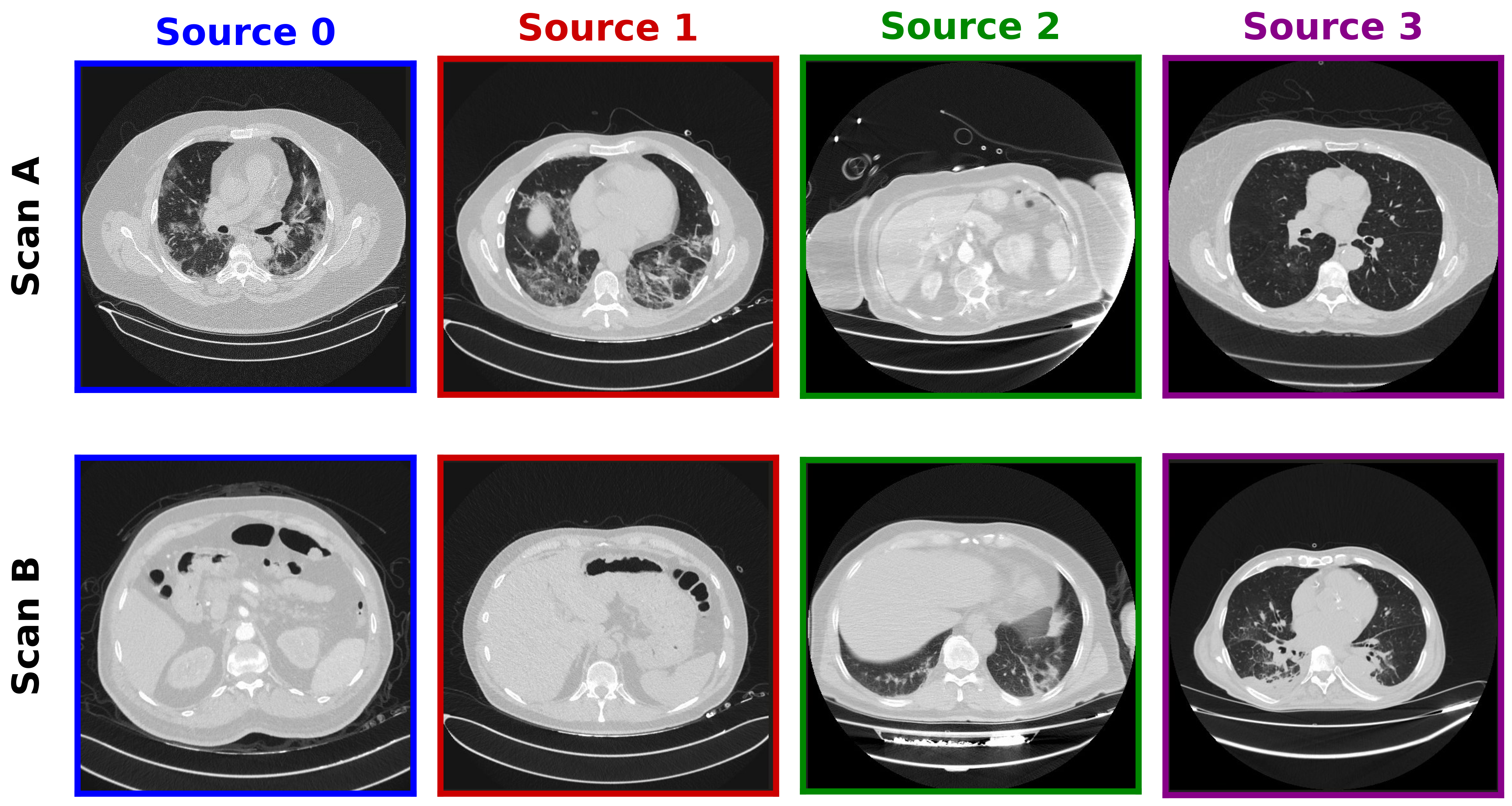}
\vspace{-0.5em}
\captionof{figure}{Representative chest CT slices from two different scans across the four contributing
medical centres, illustrating differences in field of view, contrast, tissue brightness, and windowing range that constitute the domain shift a multi-source model must overcome.}
\vspace{-5mm}
\label{fig:intro_sources}
\end{figure}

The COVID-19 pandemic, caused by the SARS-CoV-2 virus, has resulted in millions of deaths and
placed enormous strain on healthcare systems around the world. Rapid diagnosis is crucial for
managing infected patients and preventing further transmission. COVID-19 detection often relies on analyzing  chest CT images, since they can reveal lung abnormalities such as ground-glass
opacities, consolidation, and bilateral infiltrates that are indicative of the disease~\cite{kollias21,kollias22}.
However, the reality in most hospitals is that CT scans are being produced faster than radiologists can
review them. This gap between supply and capacity has driven the push toward automated screening
using deep learning~\cite{arsenos22,hou21,li25, bansal2025robust, liu2025medchat, hu2025improving, yang2024llm, huang2025diffusion, huang2024robustly, zheng2024contextual, tsai2024uu2, yang2024explainable, tsai2024uu, lin2024robust, zhu2024cgd, wang2024challenge, hu2024umednerf, wang2024neural, hu2023attention}.

The majority of existing detection models are built under a fairly narrow assumption:  \textit{the
training data and the test data come from the same hospital or at least from scanners with similar
characteristics}. This assumption breaks down quickly in multi-institutional settings. Hospitals acquire
CT scans under different protocols, using different machines, at different slice thicknesses, from
patient populations that may differ considerably in age, ethnicity, and disease severity~\cite{kollias23a}.
All of these factors introduce what is referred to as \textit{domain shift} (Figure~\ref{fig:intro_sources}), and it can significantly hurt a
model's performance when it encounters data from a centre it was not trained on~\cite{kollias23b,kollias24,kollias2024domain}.

There is also a challenge about data volume. Not all centres contribute equally. In our dataset, three
of the four sources provide around 330 training scans each, while the fourth contributes only 234. A
standard training setup does not distinguish between these sources, so the model ends up being shaped
primarily by the centres that contribute the most data.

Several methods have been proposed to address aspects of these challenges. Hsu et al.~\cite{hsu24}
introduced SSFL with Kernel-Density-based Slice Sampling (KDS), a strategy that selects a fixed number
of representative slices from each scan, making the input less tied to any particular scanner's
characteristics. Li et al.~\cite{li25} worked with full 3D CT volumes, removing slices that do
not contain lung tissue and applying a weighted cross-entropy loss to account for class imbalance. Other
studies have compared CNN architectures against transformers to understand which generalises better
across sources~\cite{hsu24}. These are useful contributions, but they all optimise for a single
objective---COVID versus non-COVID---without ever requiring the model to be aware of where each scan
was acquired. The result is that the learned features can become implicitly biased toward the dominant
sources in the training set, and this bias  becomes apparent when the model is evaluated per source.

Our work takes a step in a different direction. We retain the EfficientNet-B7
backbone~\cite{tan19} and the SSFL+KDS preprocessing from prior work~\cite{hsu24}, but we
introduce a second classification head trained to predict which of the four data centres a scan
originates from. The rationale is that by learning to solve both tasks simultaneously, the shared feature
extractor can extract feature representations that are irrelevant across all centres, not just those with the
largest presence in the training set. Because the source distribution is not uniform, we apply a
logit-adjusted cross-entropy loss~\cite{menon21} for this auxiliary head.
The logit adjustment adds an offset based on the log-frequency of each source, which prevents the
model from being dominated by the most common centre. 
Our contributions are as follows:
\begin{enumerate}
    \item We cast multi-source COVID-19 detection as a multi-task problem by pairing the diagnostic
    objective with source identification, which pushes the shared backbone toward learning features
    that transfer across institutional boundaries.

    \item We show that logit-adjusted source supervision improves the final score by 1.86 percentage points over the single-task baseline and 1.98 percentage points over the best standard multi-task variant.


\end{enumerate}

\section{Related Work}
\subsection{COVID-19 Detection}
 
Chest CT has emerged as a frontline modality for COVID-19 diagnosis, and deep learning methods have advanced rapidly in this space. Kollias et al.~\cite{kollias2022miacov19d} established an early benchmark with a large-scale 3D CT database, demonstrating that dedicated architectures can jointly detect infection and grade severity. Hou et al.~\cite{hou2021cmc} introduced contrastive mixup to capture inter-sample structure, and their later work~\cite{hou2021periphery} focused on peripheral ground-glass opacities that conventional models tend to overlook. A notable departure from volumetric processing came from Hsu et al.~\cite{hsu2024ssfl}, whose SSFL framework with Kernel-Density-based Slice Sampling showed that a handful of well-chosen 2D slices can rival full 3D pipelines, an important finding given the computational constraints of clinical deployment. Li et al.~\cite{li2025advancing} pursued the 3D route with ResNeSt50~\cite{zhang2022resnest} and lung region cropping, pairing it with weighted cross-entropy to manage class imbalance in multi-category diagnosis.
 
A fundamental shortcoming of these methods is that they treat the training distribution as representative of deployment conditions. In multi-institutional settings, this assumption breaks down: scanner vendors, reconstruction kernels, slice thickness, and patient demographics all vary across sites~\cite{kollias2023harmonizing}, and even small distributional shifts can cause substantial drops in sensitivity. Kollias et al.~\cite{kollias2023aienabled} recognized this and called for screening pipelines that explicitly integrate domain adaptation and fairness, but the question of \textit{how} to do so without access to labeled target data remains open. Yuan et al.~\cite{yuan2024pseudolabel} attempted pseudo-label-based adaptation, which improves cross-domain transfer in favorable conditions but is inherently fragile: when the source model is poorly calibrated on the target domain, pseudo-labels amplify rather than correct the distributional mismatch. Hu et al.~\cite{hu2021disentangle} offered a more principled alternative through disentanglement, separating domain-specific artifacts from generalizable features. While effective for pairwise source--target transfer, disentanglement alone does not scale naturally to the multi-source setting where the model must remain invariant to \textit{several} distinct domains simultaneously. Our framework addresses this gap directly. Rather than adapting to a specific target, we introduce source identification as an auxiliary classification task during training. This forces the shared encoder to internalize what distinguishes sources from one another, and in doing so, learn representations where diagnostic signal is preserved but institutional confounders are suppressed.

\begin{figure*}[htbp]
    \centering
    \includegraphics[width=\textwidth]{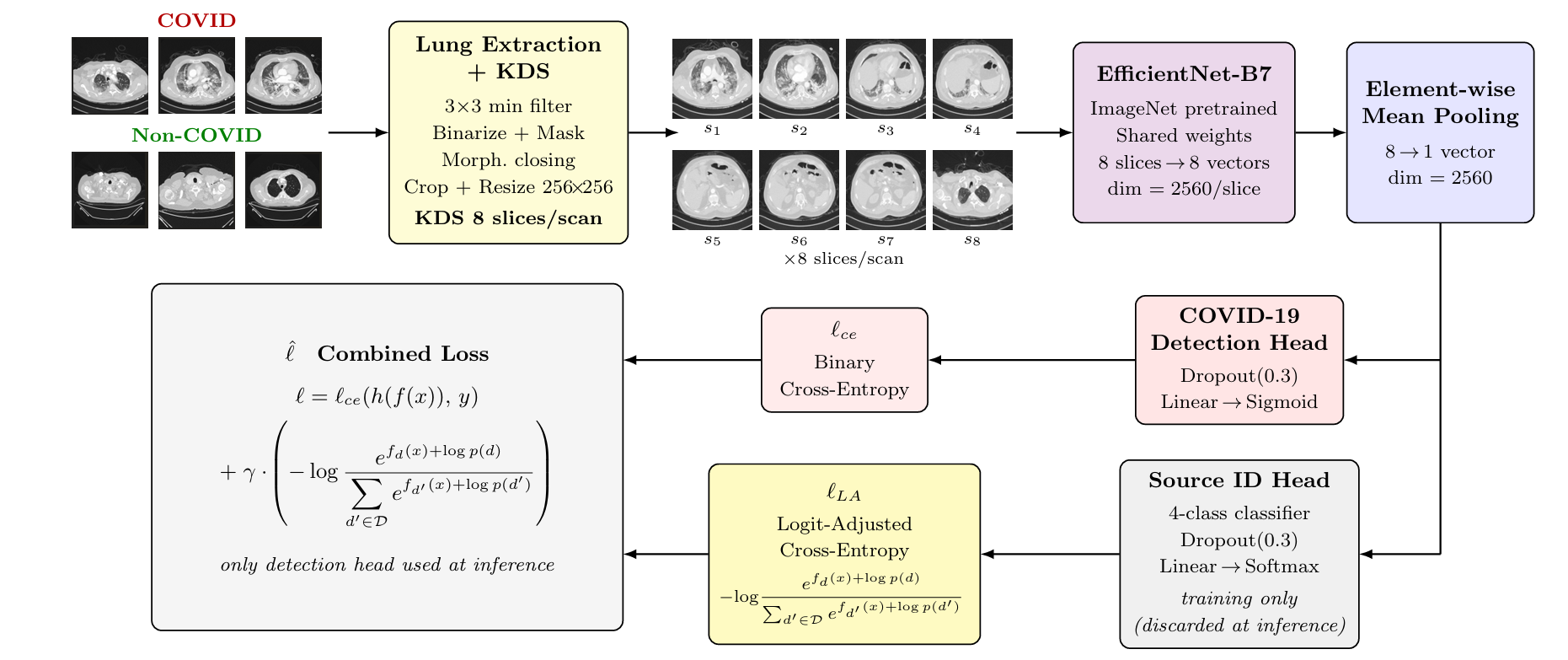}
   \caption{Overview of the proposed pipeline. Raw CT scans from COVID and non-COVID cases undergo lung extraction and KDS sampling to yield eight slices per scan. These pass through a shared EfficientNet-B7 backbone and are aggregated via element-wise mean pooling into a single 2560-dimensional scan-level representation. Two task-specific heads operate on this vector. The COVID-19 detection head is trained with binary cross-entropy ($\ell_{ce}$) and the auxiliary source identification head with logit-adjusted cross-entropy ($\ell_{LA}$). The combined objective ${\ell} = \ell_{ce} + \gamma \cdot \ell_{LA}$ governs training, while only the detection head is used at inference.}
    \label{fig:pipeline}
\end{figure*}
 
\subsection{Fairness and Class Imbalance}
 
The assumption that aggregate accuracy reflects clinical utility has come under serious scrutiny. Obermeyer et al.~\cite{obermeyer2019racial} revealed that a commercial risk-prediction algorithm systematically underestimated the health needs of Black patients, not because the model was poorly trained, but because the optimization objective itself encoded structural bias. In medical imaging, the problem manifests differently but no less consequentially: Seyyed-Kalantari et al.~\cite{seyyedkalantari2021underdiagnosis} demonstrated that chest radiograph classifiers underdiagnose pathologies more frequently in underserved populations, even when overall AUC appears strong. Kollias et al.~\cite{kollias2024fairness} confirmed that this pattern extends to COVID-19 CT, where models trained without fairness constraints exhibit pronounced accuracy gaps across demographic subgroups.
 
Addressing this requires rethinking the loss function, not just the data. Ju et al.~\cite{ju2024fairness} formalized this through CVaR-based objectives that shift optimization pressure toward worst-performing subgroups rather than the population average. Hu and Chen~\cite{hu2022dro} went further by removing the dependency on demographic annotations entirely, proposing a distributionally robust loss that achieves fairness guarantees without knowing group membership at training time, a critical advantage in clinical datasets where such labels are often unavailable or unreliable. Lin et al.~\cite{lin2024fairnessgen} tackled the harder problem of \textit{maintaining} fairness under domain shift, combining feature disentanglement with loss landscape flattening to prevent the model from converging to sharp minima that generalize unevenly across groups. These methods originate in face forensics, but the core insight, that standard ERM produces models whose failures concentrate on minority subgroups, is domain-agnostic and applies directly to multi-source medical imaging.
 
Class imbalance compounds the fairness problem \cite{ding2025fairness, hou2025rethinking, krubha2025robust, wu2025preserving, lin2025ai, lin2024robust, lin2024robust1, lin2024preserving, hu2024fairness, hu2023rank, pu2022learning, guo2022robust, hu2022sum, hu2020learning}. When certain sources contribute disproportionately to the training set, the model's decision boundaries skew toward the majority source, and underrepresented institutions effectively become out-of-distribution at test time. Cao et al.~\cite{cao2019ldam} addressed label-level imbalance through distribution-aware margins, and Hu et al.~\cite{hu2023rankbased} provided a unifying theoretical lens by showing that top-k, CVaR, and ranked-range losses are all instances of a broader rank-based decomposable framework, a perspective that clarifies when and why different rebalancing strategies succeed or fail. Menon et al.~\cite{menon21} offered perhaps the simplest effective solution: logit adjustment, where log class-frequency offsets are added to the logits before softmax. This yields Fisher consistency for the balanced error, a guarantee that neither LDAM~\cite{cao2019ldam} nor equalization loss~\cite{tan2020equalization} can provide. Despite its theoretical appeal and strong empirical results on natural image benchmarks, logit adjustment has not been applied as an auxiliary loss for source-level rebalancing in medical imaging. In our pipeline, we apply it to the source classification head, correcting for the uneven hospital-level contributions to the training set and ensuring that no single source dominates the learned decision boundaries.

\section{Method}
Our framework contains three components. First, a preprocessing pipeline that standardizes the heterogeneous inputs which inevitably arise when pooling data from four different hospitals. Second, a dual-head multi-task architecture that jointly performs disease detection and source identification, preventing the learned features from quietly overfitting to source-specific quirks. Third, a logit-adjusted training objective that corrects for unequal hospital contributions to the training set. Figure~\ref{fig:pipeline} provides an overview of the developed framework.

\begin{figure*}[t]
\centering
\includegraphics[width=\textwidth]{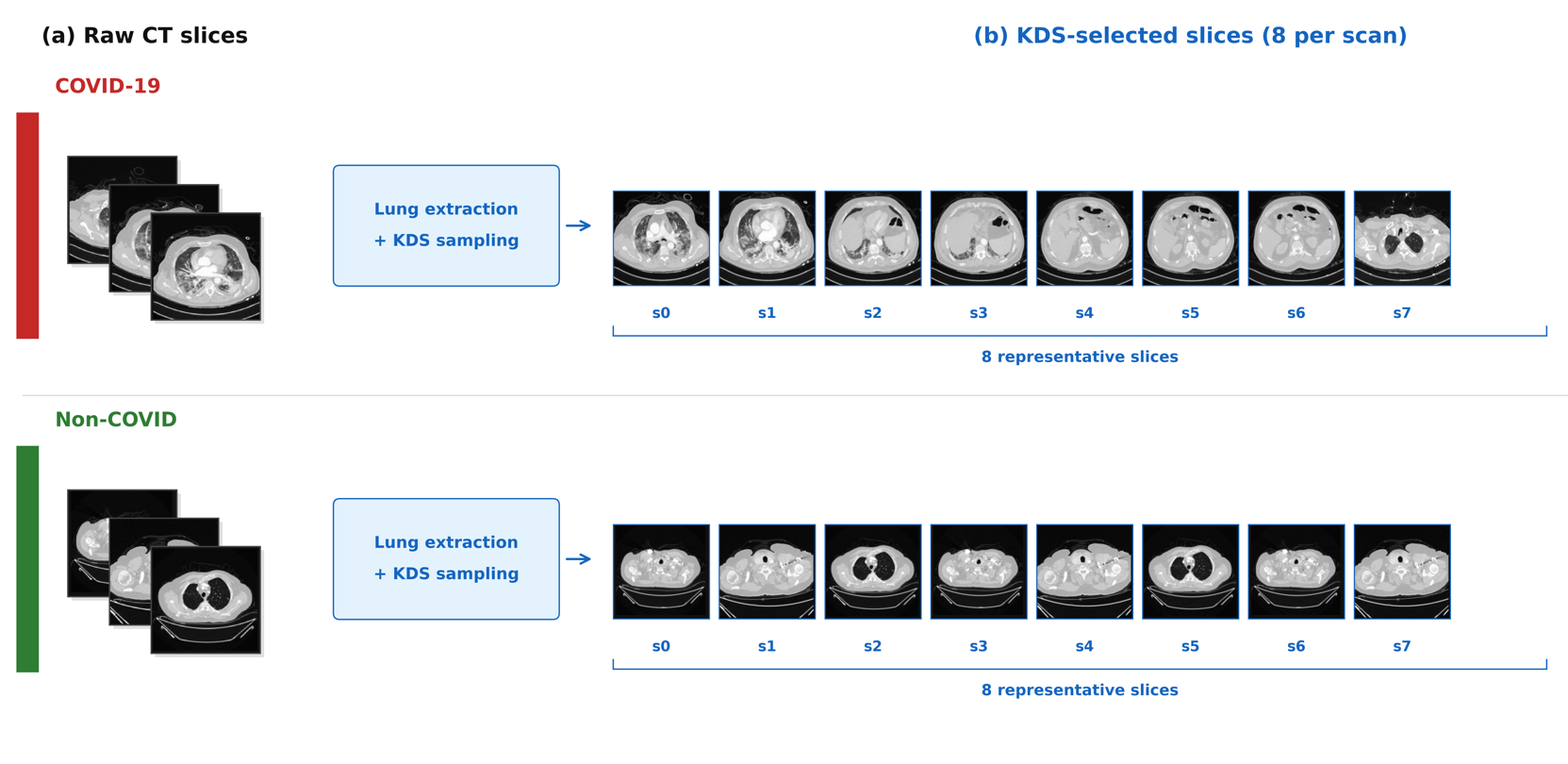}
\vspace{-2em}
\caption{Data preprocessing pipeline. Raw CT scans for both COVID-19 and Non-COVID cases undergo lung extraction and KDS sampling to select 8 highly representative slices per scan.}
\vspace{-5mm}
\label{fig:preprocessing}
\end{figure*}
\vspace{1em}

\subsection{Preprocessing}
Multi-centre CT data presents a familiar but persistent problem. No two hospitals acquire scans the same way. Scanner hardware differs, reconstruction kernels differ, slice thickness differs, and patient positioning varies across sites. If we feed this raw heterogeneity directly into a classifier, there is nothing stopping the model from latching onto acquisition artifacts rather than genuine pathological features. We address this through a two-stage pipeline.

The first stage isolates the lung region in each 2D slice. A $3\times3$ minimum filter suppresses high-frequency noise and scanner-specific artifacts. The filtered image is binarized at a fixed intensity threshold, and morphological hole filling produces a clean binary lung mask. Slices that fail basic consistency checks, dimension mismatches within a scan folder or folders containing fewer than five usable images, are discarded. The remaining lung region is cropped and resized to $256\times256$ pixels, establishing uniform spatial resolution across all four centres regardless of the original acquisition protocol~\cite{kollias21}.

The second stage tackles a subtler issue. Scan lengths vary enormously across sources. One hospital might produce 50 slices per patient while another produces 300. Naive uniform subsampling would oversample long scans and undersample short ones, and in doing so risk missing the most diagnostically informative regions entirely. To handle this, we employ Kernel Density based Slice Sampling (KDS)~\cite{kollias22}. The lung area of every valid slice in a scan is computed, and a Gaussian kernel density estimate is fitted over that distribution with bandwidth chosen by Scott's rule. The resulting cumulative distribution function is partitioned into eight equal percentile intervals, and from each interval the slice nearest to the midpoint is selected. This yields exactly eight slices per scan regardless of the original length  (Figure~\ref{fig:preprocessing}). For the occasional cases where very short scans produce duplicate selections, we pad them by repeating the last chosen slice.

What makes this scheme worth the added complexity over uniform sampling is that it naturally adapts to the underlying anatomy. Slices where the lung cross-section changes rapidly, typically around the hilum and the bases, end up more densely represented than the relatively uniform apical regions. The eight selected frames therefore amount to a compact but anatomically faithful summary of the entire volume, one that captures the regions most likely to carry diagnostic signal without wasting representation on redundant slices.

\subsection{Multi-Task Model Architecture}
We select B7 specifically rather than a smaller variant in the EfficientNet family~\cite{tan19}. Among the available configurations, B7 offers the most aggressive compound scaling of depth, width, and input resolution, which becomes important when the visual differences between COVID-positive and COVID-negative CT slices can be exceedingly subtle. We initialize from ImageNet pretrained weights, not because natural image features are directly relevant to lung pathology, but because the low-level filters for edges, textures, and local contrast that ImageNet pretraining produces give the network a substantial head start compared to random initialization. This is well-established in the medical imaging literature and we found it to hold in our setting as well.

Each scan's eight preprocessed slices pass independently through this shared backbone, producing eight feature vectors of dimension 2560. We aggregate them via element-wise mean pooling into a single scan-level representation. We considered more expressive alternatives. Attention pooling could in principle learn to weight certain slices more heavily. A recurrent layer could exploit ordering. But KDS selects slices by percentile bin rather than by anatomical position, so there is no meaningful sequential structure to exploit, and attention risks overfitting to spurious slice-level patterns in a training set of roughly 1200 scans. Mean pooling treats every selected slice as an equally valid window into the scan, which aligns with the design philosophy behind KDS and the SSFL framework~\cite{hsu24} and avoids introducing unnecessary parameters.

Two task-specific classification heads operate on this shared 2560-dimensional vector. The COVID-19 detection head applies dropout at rate 0.3 followed by a single linear unit with sigmoid activation, producing the probability that the scan is positive. The source identification head has the same dropout but terminates in a four-way linear layer with softmax, predicting which of the four hospitals produced the scan.

The inclusion of source prediction as an auxiliary task warrants justification, since the source label is not relevant at inference and the additional head introduces both parameters and a competing gradient signal. The motivation is rooted in a specific failure mode of single-task training. If we optimize only for COVID-19 detection, the backbone is free to learn whatever features minimize that single loss, and nothing prevents those features from encoding hospital identity rather than disease pathology. A reconstruction kernel unique to one centre, a brightness distribution characteristic of another. These are easy signals for a deep network to exploit and they generalize poorly. By simultaneously requiring the network to classify the source, we force the shared representation to disentangle disease signal from acquisition signal. The source head's only job is to make itself unnecessary, to push the backbone toward features that would work equally well no matter which hospital the scan came from. At inference, the source head is discarded entirely. It never touches the final prediction.

\subsection{Logit-Adjusted Source Classification Loss}
There is a practical complication. Three of the four centres each contribute roughly 330 scans to the training set, but the fourth contributes only 234. This will cause the source head gradually drifts toward overpredicting the majority centres under learning with standard cross-entropy loss, simply because doing so is the path of least resistance for minimizing the loss. And because gradients from the source head flow directly back into the shared backbone, this bias does not stay contained. It seeps into the feature space itself, which is exactly the opposite of what the auxiliary task is supposed to achieve.

We correct for this with logit-adjusted cross-entropy following Menon et al.~\cite{menon21}. Given the predicted source logit $f_d(x)$ for an input $x$ and the true source domain $d \in \mathcal{D}$, where $\mathcal{D}$ is a source domain set, the adjusted loss adds a correction term derived from the empirical class frequencies:

\begin{equation}
    \ell_{\text{LA}} = -\log \left( \frac{e^{f_d(x) + \log(p(d))}}{\sum_{d' \in \mathcal{D}} e^{f_{d'}(x) + \log(p(d'))}} \right)
\end{equation}

where $p(d)$ is the source proportion (prior probability) computed from the training set. The mechanism is simple but effective. Adding $\log(p(d))$ to the logit for centre $d$ before the softmax shifts the decision boundary so that predicting a rare source demands less raw confidence than predicting a common one. The model can no longer coast by defaulting to the majority class. Underrepresented centres keep receiving meaningful gradient signal throughout training, which is exactly the condition needed for the shared backbone to develop a genuinely source-invariant feature space rather than one biased toward whichever hospitals happened to contribute more data.

\subsection{Learning Objective}
The total loss combines both heads through a weighted sum.

\begin{equation}
    \ell_{\text{}} = \ell_{\text{ce}}(h(f(x)), y) + \gamma \left( -\log \left( \frac{e^{f_d(x) + \log(p(d))}}{\sum_{d' \in \mathcal{D}} e^{f_{d'}(x) + \log(p(d'))}} \right) \right)
\end{equation}

Here, $l_{ce}(h(f(x)), y)$ is the standard binary cross-entropy applied to the COVID-19 detection head for an input with true label $y$. $\gamma \geq 0$ controls how much influence the source task exerts during backpropagation. When $\gamma = 0$, the source head receives no gradient and the model collapses to a plain single-task classifier. As $\gamma$ grows, the backbone is pulled progressively harder toward representations that must remain consistent across all four hospitals.

This creates a tension that is worth being explicit about. Too small a $\gamma$ and the source regularization is toothless, present in the computational graph but too weak to meaningfully shape the features. Too large and the auxiliary objective overwhelms the primary detection task. In the worst case, the backbone starts encoding source identity so aggressively that disease features get crowded out, which defeats the entire purpose.


\section{Experiments}
\subsection{Experimental Settings}
\textbf{Datasets}.
All experiments are conducted on the Multi-Source COVID-19 Detection Database~\cite{kollias2025pharos}, which aggregates 3D chest CT scans from four institutionally distinct medical centres, identified as Sources 0 through 3. The database contains 3,020 scans in total, comprising 1,035 COVID-19 positive and 1,985 negative cases, each manually annotated and anonymised. Individual scans vary considerably in depth, ranging from as few as 50 to as many as 700 axial slices, though the in-plane resolution is fixed at $512 \times 512$ pixels throughout. After applying our SSFL preprocessing pipeline (Section~3.1), which discards scans containing fewer than five valid slices, 1,222 scans remain for training and 308 for validation. The per-source breakdown is given in Table~\ref{tab:dataset}.

\begin{table}[H]
\centering
\caption{Per-source distribution of scans in the training and validation sets. Source 3 contains no COVID-19 cases in the validation partition.}
\label{tab:dataset}
\resizebox{\columnwidth}{!}{%
\begin{tabular}{c|ccc|ccc}
\toprule
 & \multicolumn{3}{c|}{\textbf{Training}} & \multicolumn{3}{c}{\textbf{Validation}} \\
\textbf{Source} & \textbf{COVID} & \textbf{Non-COVID} & \textbf{Total} & \textbf{COVID} & \textbf{Non-COVID} & \textbf{Total} \\
\midrule
0 & 165 & 163 & 328 & 45 & 45 & 90 \\
1 & 165 & 165 & 330 & 45 & 45 & 90 \\
2 & 165 & 165 & 330 & 38 & 45 & 83 \\
3 & 69 & 165 & 234 & 0 & 45 & 45 \\
\midrule
\textbf{Total} & \textbf{564} & \textbf{658} & \textbf{1,222} & \textbf{128} & \textbf{180} & \textbf{308} \\
\bottomrule
\end{tabular}%
}
\end{table}

The class distribution exhibits a pronounced asymmetry across sources. Sources 0, 1, and 2 are roughly balanced, each contributing approximately 165 COVID-19 training cases and 38--45 positive validation cases. Source 3, however, deviates substantially: it provides only 69 COVID-19 cases for training, less than half the count of the other centres, and contributes zero positive cases to the validation set. As a consequence, the per-source COVID-19 F1 for Source 3 defaults to 0.0 for every method we tested, placing a hard structural ceiling on the achievable final score. The non-COVID-19 distribution, by contrast, is relatively uniform across all four sources at roughly 163--165 training scans each. No external data is used in any of our experiments.
 
\smallskip
\noindent
\textbf{Evaluation Metrics}.
We use F1 score to assess binary classification quality, as it balances precision and recall of the model's diagnostic predictions, a standard choice in COVID-19 detection literature~\cite{kollias21,hsu24}. We also report AUC-ROC to capture discriminative ability across the full range of decision thresholds, independent of any particular operating point~\cite{kollias22}.
 
Our primary evaluation criterion, however, is the official competition final score, which is specifically designed to expose cross-centre inconsistency. For each source $i \in \mathcal{D}$, the F1 is computed separately for the COVID-19 and non-COVID-19 classes, and the two are averaged. The final score is then the mean of these per-source averages:
\begin{equation}
\text{Final Score} = \frac{1}{|\mathcal{D}|}\sum_{i \in \mathcal{D}} \frac{F1_{\text{covid}}^{(i)} + F1_{\text{non-covid}}^{(i)}}{2}
\end{equation}
 
Crucially, this formulation is not dominated by majority sources: strong performance on well-represented centres cannot offset degradation elsewhere. A model that achieves near-perfect classification on Sources 0--2 but collapses on Source 3 will still receive a low final score, directly incentivising representations that generalise across institutional boundaries.

\smallskip
\noindent
\textbf{Baseline Methods}. 
We define two baselines to cleanly attribute performance gains to specific design choices. Both use the same backbone architecture, the same preprocessing pipeline (SSFL lung extraction followed by KDS slice sampling, as described in Section~3.1), the same augmentation strategy, and the same optimiser configuration detailed in Section~4.1.4. The only thing that changes is the loss function.
 
The first is a \textit{single-task baseline}. Here, the COVID-19 classification head is trained with binary cross-entropy alone; no source classifier is attached, and no multi-source signal is exploited. In our formulation this corresponds to $\gamma = 0$. It serves as the standard transfer-learning reference: a strong pretrained backbone fine-tuned on the target task with no auxiliary supervision.
 
The second is a \textit{multi-task CE baseline}. This variant adds a four-class source classification head and trains it jointly with the COVID-19 head. The source branch is supervised with standard cross-entropy, without any frequency-based correction, yielding a combined objective of $\mathcal{L}_{\text{BCE}} + \gamma \cdot \mathcal{L}_{\text{CE}}$. We sweep $\gamma$ over the same grid as our proposed method.
 
This design disentangles two factors that would otherwise be confounded. The single-task versus multi-task comparison isolates the effect of auxiliary source supervision itself, while the CE versus LA comparison, with all other variables held constant, directly quantifies the contribution of frequency-aware logit adjustment.
 
\smallskip
\noindent
\textbf{Implementation Details}. 
We use EfficientNet-B7~\cite{tan19} as the backbone, loaded with ImageNet pretrained weights from the \texttt{timm} library (\texttt{tf\_efficientnet\_b7}). The network produces a 2,560-dimensional feature vector per input image. Each CT scan is first reduced to eight representative slices via KDS (Section~3.1). These slices are forwarded through the backbone independently, and the resulting eight feature vectors are mean-pooled to produce a single scan-level representation. The resulting scan-level vector is forwarded to two classification heads: a binary COVID-19 head producing a single logit, and a four-class source head. Both heads are architecturally identical, consisting of a dropout layer ($p = 0.3$) followed by a linear projection, differing only in output dimensionality.
 
During training, we apply a suite of per-slice spatial and photometric augmentations via the Albumentations library~\cite{buslaev2020}: random horizontal flipping ($p = 0.5$), shift-scale-rotate with shift limit 0.2, scale limit 0.2, and rotation up to $30^\circ$ ($p = 0.5$), hue-saturation-value jitter with a limit of 20 on each channel ($p = 0.5$), random brightness and contrast adjustment within $\pm 0.2$ ($p = 0.5$), and coarse dropout that masks up to 8 rectangular regions of $32 \times 32$ pixels ($p = 0.2$). All slices are resized to $256 \times 256$ and normalised using ImageNet statistics. At validation time we apply only resizing and normalisation; no test-time augmentation is used.
 
We train with the Adam optimiser~\cite{kingma2015adam} using a learning rate of $1 \times 10^{-4}$ and weight decay of $5 \times 10^{-4}$, a batch size of 10, and mixed-precision arithmetic via PyTorch AMP. The multi-task weighting hyperparameter $\gamma$ is swept over $\{0.1, 0.2, 0.5, 1.0\}$. Each configuration is trained for 8 epochs, and we retain the checkpoint that achieves the highest validation F1. All experiments are conducted on a single NVIDIA A100 GPU using PyTorch.

\subsection{Results}


\textbf{Overall Performance.} Table~\ref{tab:results} summarises validation and final score results. The single-task baseline achieves a final score of 0.8008 (F1 = 0.8915, AUC = 0.9627). Adding the source head with standard CE does not improve on this. The best CE configuration scores 0.7996 at $\gamma = 0.1$, and stays below the baseline at every $\gamma$ tested. This points to a gradient conflict. Without any correction for uneven sample counts across centres, the CE source head biases shared features toward majority-centre statistics, and the primary task suffers as a result.

Replacing CE with logit-adjusted cross-entropy resolves this. At $\gamma = 0.5$, the final score rises to \textbf{0.8194}, up 1.86 points over the baseline and 1.98 over the best CE variant. This run also produces the highest F1 (0.9098), correctly identifying 169/180 non-COVID scans (specificity = 0.9389) and 116/128 COVID scans (sensitivity = 0.9062). The idea is simple. Subtracting log-prior frequencies from the source logits~\cite{menon21} before the loss computation down-weights gradients from over-represented centres. This discourages the backbone from latching onto acquisition-specific patterns and pushes it toward features that transfer across centres, acting as implicit domain regularisation without any explicit domain-alignment objective.

\begin{table*}[htbp]
\centering
\caption{Validation results across all configurations. Baseline = BCE only, no source head. \textbf{Bold} = best per column. $\uparrow$ = higher is better.}
\label{tab:results}
\begin{tabular}{l c c c c c c c}
\hline
Configuration & $\gamma$ & Accuracy & F1 $\uparrow$ & AUC $\uparrow$ & Sensitivity & Specificity & Final Score $\uparrow$ \\
\hline
Baseline (BCE) & -- & 0.9091 & 0.8915 & 0.9627 & 0.8984 & 0.9167 & 0.8008 \\
Multi-task (CE) & 0.1 & 0.9058 & 0.8889 & \textbf{0.9683} & 0.9062 & 0.9056 & 0.7996 \\
Multi-task (CE) & 0.2 & 0.9026 & 0.8846 & 0.9486 & 0.8984 & 0.9056 & 0.7923 \\
Multi-task (CE) & 0.5 & 0.8896 & 0.8692 & 0.9523 & 0.8828 & 0.8944 & 0.7904 \\
Multi-task (CE) & 1.0 & 0.9058 & 0.8930 & 0.9715 & \textbf{0.9453} & 0.8778 & 0.7942 \\
Multi-task + LA & 0.1 & 0.9123 & 0.8924 & 0.9675 & 0.8750 & \textbf{0.9389} & 0.7986 \\
Multi-task + LA & 0.2 & 0.8994 & 0.8794 & 0.9561 & 0.8828 & 0.9111 & 0.7850 \\
Multi-task + LA & 0.5 & \textbf{0.9253} & \textbf{0.9098} & 0.9647 & 0.9062 & \textbf{0.9389} & \textbf{0.8194} \\
Multi-task + LA & 1.0 & 0.8929 & 0.8800 & 0.9462 & \textbf{0.9453} & 0.8556 & 0.7910 \\
\hline
\end{tabular}
\end{table*}

Worth noting is multi-task CE at $\gamma = 1.0$. It achieves F1 = 0.8930 and the highest AUC of any run (0.9715), yet its final score sits at just 0.7942. High aggregate accuracy can mask poor per-source balance. The model concentrates capacity on data-rich centres, and the equally-weighted final score exposes exactly that. The LA loss targets this failure mode by flattening gradient contributions across sources.

\begin{figure}[!htbp]
\centering
\includegraphics[width=\columnwidth]{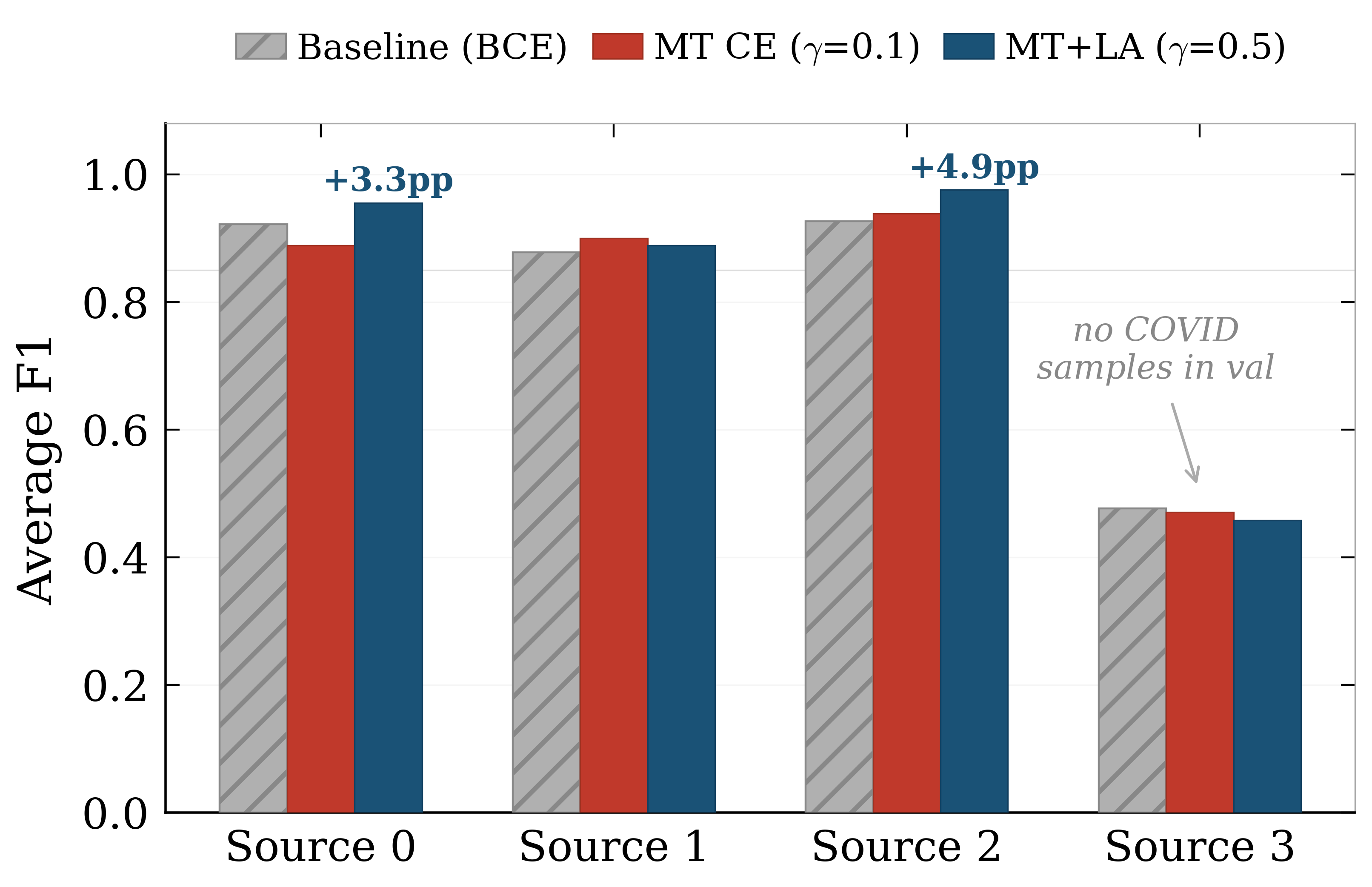}
\caption{Per-source average F1 at each method's best $\gamma$. The LA variant yields the largest gains on Source~0 (+3.3\,percentage point) and Source~2 (+4.9\,percentage point). Source~3 scores remain low across all methods due to the absence of COVID samples in the validation split.}
\label{fig:persource}
\end{figure}

\smallskip
\noindent
\textbf{Per-Source Analysis.} Table~\ref{tab:persource}  and Figure~\ref{fig:persource} break down F1 by source at each method's best $\gamma$. The final score~\cite{kollias2025pharos} weights all four centres equally, computed as $\frac{1}{4}\sum_{i=0}^{3}\frac{F1_{\text{covid}}^i + F1_{\text{non-covid}}^i}{2}$.

\begin{table}[htbp]
\centering
\caption{Per-source average F1 at each method's best $\gamma$. \textbf{Bold} = best per source.}
\label{tab:persource}
\small
\setlength{\tabcolsep}{3pt}
\begin{tabular}{l c c c c c}
\hline
Method & Src 0 & Src 1 & Src 2 & Src 3 & Final $\uparrow$ \\
\hline
Baseline (BCE) & 0.9221 & 0.8776 & 0.9269 & 0.4767 & 0.8008 \\
MT CE ($\gamma$=0.1) & 0.8888 & \textbf{0.9000} & 0.9389 & 0.4706 & 0.7996 \\
MT + LA ($\gamma$=0.5) & \textbf{0.9555} & 0.8888 & \textbf{0.9756} & 0.4578 & \textbf{0.8194} \\
\hline
\end{tabular}
\end{table}     

The gains concentrate on the centres where the baseline leaves the most room for improvement. Source~2 improves by 4.87 percentage points over the baseline to 0.9756, and Source~0 by 3.34 percentage points. Notably, the training-set class imbalance lies in Source~3 (69 COVID vs.~165 non-COVID), not Source~2, which is nearly balanced (Table~\ref{tab:dataset}). These improvements therefore reflect the logit adjustment redistributing gradient contributions across centres rather than correcting within-source class skew.

Source 1 dips from 0.9000 under CE to 0.8888 with LA, though it stays above the baseline (0.8776). Some redistribution of capacity away from already well-served centres is the expected cost of rebalancing. The net effect on the final score is clearly positive.

Source 3 averages below 0.50 everywhere, but this reflects the validation split, not model quality. All 45 Source 3 scans are non-COVID, so COVID F1 enters the average as zero. Non-COVID F1 for this centre is 0.9888 at $\gamma = 0.1$ (multi-task + LA) and 0.9535 under the baseline. The model handles Source 3 well on what data is available.

\subsection{Sensitivity Analysis}
\begin{figure}[!htbp]
  \centering
  \includegraphics[width=0.85\linewidth]{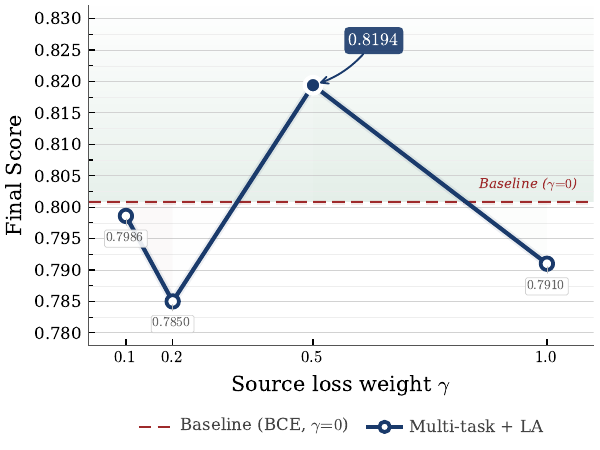}
  \caption{%
    \textbf{Sensitivity to the source-loss weight~$\gamma$.}
    Final score of \emph{Multi-task + LA} as $\gamma$ varies in
    $\{0.1,\,0.2,\,0.5,\,1.0\}$.
    The dashed line marks the BCE-only baseline ($\gamma\!=\!0$).
    Performance peaks at $\gamma\!=\!0.5$ (score $=0.8194$), where the
    source head and logit adjustment reinforce each other;
    both lower and higher weights degrade the score.}
  \label{fig:gamma_sensitivity}
\end{figure}
 
\paragraph{Sensitivity to source loss weight~$\gamma$.}
We train the multi-task + LA configuration at
$\gamma\in\{0.1,\,0.2,\,0.5,\,1.0\}$ and compare against the
BCE-only baseline ($\gamma=0$);
Fig.~\ref{fig:gamma_sensitivity} reports the final score for each
setting.
The trend is non-monotonic.
At $\gamma=0.1$ the final score ($0.7986$) remains close to the
baseline of $0.8008$, indicating that a lightly weighted source head
adds little beyond what logit adjustment already provides.
At $\gamma=0.2$ performance drops to $0.7850$: the source loss is
strong enough to interfere with the detection gradient but not yet
sufficient to produce a coherent domain signal for the logit-adjusted
loss to act on.
At $\gamma=0.5$ the score rises to $\mathbf{0.8194}$, the highest
across all configurations, where the source head provides enough
per-centre contrast for logit adjustment to effectively redistribute
gradient contributions across sources, and the two objectives reinforce
each other.
At $\gamma=1.0$ this balance breaks: the source objective dominates the
shared encoder, shifting representations toward centre identity rather
than disease-discriminative features, and the score falls to $0.7910$.
We therefore fix $\gamma=0.5$ for all subsequent experiments.

\section{Conclusion}

We presented a multi-task framework for multi-source COVID-19 detection from chest CT that pairs binary diagnosis with auxiliary source identification over a shared EfficientNet-B7 backbone. The central lesson from our experiments is not that multi-task learning helps, it is that multi-task learning helps only when the auxiliary loss is designed with the data distribution in mind. A source classification head trained with standard cross-entropy inherits the same imbalance it is meant to correct, and the resulting gradients quietly steer the shared encoder toward majority centre patterns rather than away from them. Logit-adjusted cross-entropy breaks this cycle by flattening gradient contributions across sources. 

\smallskip
\noindent
\textbf{Limitations.} Our framework assumes that centre identity labels are available during training, which may not hold in fully anonymised settings. The preprocessing pipeline also applies a fixed strategy across all sources rather than adapting to source specific scan characteristics.

\smallskip
\noindent
\textbf{Future Work.} The dependency on source labels could be relaxed through unsupervised domain discovery. Incorporating patient level fairness objectives alongside source level rebalancing would provide stronger equitable performance guarantees across demographic subgroups.

\section*{Acknowledgements}
This work is supported by the U.S. National Science Foundation (NSF) under grant IIS-2434967, and the National Artificial Intelligence Research Resource (NAIRR) Pilot and TACC Lonestar6. The views, opinions and/or findings expressed are those of the author and should not be interpreted as representing the official views or policies of NSF and NAIRR Pilot.

{
    \small
    \bibliographystyle{IEEEtran}
    \bibliography{main}
}


\end{document}